# CLIC: A Framework for Distributed, On-Demand, Human-Machine Cognitive Systems

N. Mavridis (1), S. Konstantopoulos (1), I. Vetsikas (1),
I. Heldal (2), P. Karampiperis (1), G. Mathiason (2), S. Thill (2),
K. Stathis (3), V. Karkaletsis (1)

(1) Institute of Informatics and Telecommunications
NCSR "Demokritos", Athens, Greece
(2) University of Skovde, Sweden
(3) Royal Holloway, University of London, U.K.

Abstract

Traditional Artificial Cognitive Systems (for example, intelligent robots) share a number of common limitations. First, they are usually made up only of machine components; humans are only playing the role of user or supervisor. And yet, there are tasks in which the current state of the art of AI has much worse performance or is more expensive than humans: thus, it would be highly beneficial to have a systematic way of creating systems with both human and machine components, possibly with remote non-expert humans providing snippets of some seconds of their capacities in real-time. Second, their components are specific and dedicated to one and only one system, and are often underutilized for significant fractions of their lifetime. Third, there is no inherent support for robust, fault-tolerant operation, and if a new component becomes available, with better performance and/or cheaper cost, one cannot easily replace the old component. Fourth, and quite importantly in terms of their economics, they are viewed as a resource that needs to be developed and owned, not as a utility; i.e. not as a service provided on demand.

Motivated by the above state of affairs, in this paper we are presenting CLIC: a framework for constructing cognitive systems that overcome the above mentioned limitations. With the four-layer software architecture of CLIC, we provide specific yet extensible mechanisms that enable the creation and operation of distributed cognitive systems that fulfill the following desiderata: First, that are distributed yet situated, interacting with the physical world though sensing and actuation services, and that are also combining services provided by humans as well as services implemented by machines. Second, that are made up of components that are time-shared and re-usable across systems. Third, that provide increased robustness through self-repair mechanisms. Fourth, that are constructed and reconstructed on the fly, with components that dynamically enter and exit the system, while the system is in operation, on the basis of availability, and pricing, and need. Quite importantly, fifth, the cognitive systems created and operated by CLIC do not need to be owned and can be provided on demand, as a utility – thus transforming human-machine situated intelligence to a service, and opening up numerous interesting research directions and application opportunities.

# 1. Introduction

Traditionally, most of the cognitive systems that have been engineered in the past, have had a number of *common characteristics*. First, they were primarily composed of electronic or machine elements; and *not human elements*. For example, consider the case of an intelligent robot equipped with vision that aims to fulfill a mechanical task – search for a certain kind of object, fetch, and place it at a specific container. Such a system is comprised of electronic *sensing* components (camera, sonars etc.), electronic *cognition* components (software components performing pattern recognition, planning, motor control etc.), and electromechanical *actuation*: an arm with a gripper, which fetches and places objects. Notice that there are *no human component*s within such a cognitive system; even if the system was interacting with a human, the human is not part of the sensory, cognitive or actuation components of the system; but rather has the role of a supervisor, controller, or collaborator which is external to the cognitive system itself. Second, notice that the actual components of the cognitive system (in this case, a robot) that was just described are all *physically part of the system* itself; they are not spatially distributed, and furthermore, they are *specific* to this system (robot) and dedicate *all of their operating time* to this robot. I.e. they are usually neither *distant*, nor *distributed*, and neither *time-shared* nor *reused* for various cognitive systems. Furthermore, the components not only do not partake to different cognitive systems, but the cognitive system is dependent on them throughout its operating time; if a sensor fails, there is no graceful way for it to be replaced without disrupting the operation of the system. Thus, there is no inherent support for *robust*, fault-tolerant operation, and furthermore, if a new component (for example, camera or pattern recognition algorithm becomes available, with better performance and/or cheaper cost, one cannot easily *replace* the old component.

The motivation behind this paper is to contribute to improving the above-described state of affairs. We address the following questions: How can the cognitive systems of the future exhibit improved characteristics? Can they consist also of *human components*, given that for some cases they exhibit better performance and/or are more readily available than electronic elements? For example: humans are much better in performing activity recognition as compared to the state-of-the art automated systems (better performance). Another example is that of a human observer next to a broken traffic camera, who can be useful acting as a sensor that reports traffic conditions (availability). Based on this the following questions are raised up: Can the cognitive systems of the future include *distributed* components that are *time-shared* and *re-used* for various systems, and also enable much higher robustness and flexibility? For example – why should the surveillance cameras of a city be dedicated only to surveillance, and why can they not be reused for other purposes too? Can we even reach a stage, where one is able to offer *situated cognitive systems as an on-demand service*, on the basis of the type of the requests? All of this seems like quite distant from the state of the art of today – but is it?

It might not be that distant: it is worth noting a number of recent developments, from across the Atlantic: First, the DARPA 2009 Network Challenge, often referred to as the "Ten Red Balloons" competition. During this Challenge, ten large red balloons were placed in locations around the United States, with their location unknown to the participating teams. The goal of the teams was to create a system that is able to locate the balloons in minimum time. There was no restriction regarding the components of the system: they could be *electronic, human, or both*. A team from the MIT Media Lab won the challenge: through an ingenious scheme thousands of non-expert humans were recruited lending some seconds of their eyes to the resulting system; this information was propagated and combined, in order for the system to determine the location of the balloons, see Tang et al. [Tan11]. One can view the system that was created as a massive distributed cognitive system: with sensing (vision) provided by human components, pattern recognition (red balloon recognition) provided by humans too, and information fusion as well as propagation provided by electronic components. Notice that in this system the components are *human as well as machine*, they are *distant* (spread over a large geographical area). Furthermore, the human components are *not dedicated* to the system (i.e. the humans that spent 10 seconds of their time looking around for a balloon are also using their eyes and brains for other tasks), i.e. their sensing as well as cognition apparatus is *time-shared* and *re-used*. Finally, there is a large degree of *robustness* to the system as false-reports can be crossed-out of the system through the special algorithms used and through the inherent redundancy in sensing resources.

But there are also many other recent developments of a similar nature: Another notable example is Von Ahn's [Ahn08] ingenious CAPTCHA-breaker scheme. CAPTCHAs are often used in order to stop spam email programs and other bots from creating thousands of email accounts in order to propagate spam. They are usually strings of letters and numbers, with character sets that contain geometric distortions and occlusions. The characters in the CAPTCHAs are very easy for humans to recognize; however, they are quite difficult for machines, given the state of the art of Optical Character Recognition (OCR). Thus, a solution towards breaking CAPTCHAs, involves finding non-expert humans online, and incentivizing them adequately, so that they break the CAPTCHA (by recognizing the characters) with the answer collected by the spam mailer program, which opens the accounts right away. The humans effectively lend some seconds of their mind to the system, performing the cognitive service of character recognition for it, incentivized by illegal downloads that the system offers to them in return for their services. In essence, a *large distributed cognitive system* is effectively created consisting of *human as well as machine components*, which dynamically enter and exit the system, in order to be able to achieve superior results that would have been impossible by either alone as demonstrated in the DARPA 10 Red Balloons challenge.

Finally, we consider a third recent development, the *cloud computing* paradigm. Traditionally, computation required ownership of resources: computers, storage space, and software. With cloud computing, computation is viewed as a utility. What do we mean by "utility"? In a similar sense with modern power and water networks, users of the cloud do not need to own the means of production or distribution (i.e. power generators, water sources and distribution networks): they just connect to the cloud, and time-share *reusable distant distributed* computation, storage, and code resources, in a transparent fashion (not knowing the whereabouts or the specifics of them), and with high robustness.

But how are the cloud, CAPTCHA breakers, and the ten red-balloon challenge relevant to the cognitive systems of the future? To answer this question, the aim of CLIC is to provide a conceptual framework to integrate ideas, constructs, architectures and techniques from human-machine cognitive systems, artificially intelligent agents and services of the kind presented in cloud computing to build human-machine cognitive and intelligent systems on demand. The contribution of the proposed framework is the identification and conceptual definition of four layers that need to be available for building cognitive systems applications. More concretely, the CLIC framework specifies the components in each layer, the relations between components within a layer, as well as the relation of components among different layers. The ideas presented are also exemplified using a cognitive system application for a transport system so that the reader can appreciate the broader implications and relevance of CLIC for the construction of future cognitive systems.

The paper is structured as follows. We start by providing an overview of the CLIC framework and a description of its basic concepts and terminology as well as the proposed architecture. Then we describe a use case built by the CLIC framework: a cognitive system for transport management. An extensive discussion and juxtaposition of how the four layers of our framework can be applicable to a practical application follows. We continue with a detailed discussion that identifies the research challenges for developing systems in the CLIC framework, while at the same time we review existing research that is relevant to the ideas of the framework and how such research opens up opportunities for further work and real-world large-scale applications.

## 2. Overview and Basic Concepts

The basic entity of CLIC is that of a *cognitive system* which is construed conceptually as a (possibly large) *monolithic agent*; composed of *components* that partake in an *information processing topology*. In this paper we will discuss use cases that construct feed-forward pipelines, the simplest topology possible, although arbitrarily complex topologies are admissible by the framework.

The components themselves (for example, a human sensing component) can belong to other agents (if we also consider humans themselves as agents). Since cognitive systems can be constructed as having an underlying multi-agent network, we also can adopt the viewpoint of the *composite unitary agent* that is created from them, and not that of the underlying multi-agent network.

In order to address the design challenge of distributed, human-machine, on demand, and on-the-fly cognitive systems, we propose a two-fold extension of the concept of *cloud computing* and we develop the mechanisms that enable the construction and deployment of cognitive systems using our proposed framework:

- Cloud computing is a computing paradigm of interchangeable processing and storage components that are *distributed, time-shared,* and *re-used.*
- Cloud computing is also well suited for defining *fault-tolerance* mechanisms for falling back to alternative processing and storage services.
- The first extension pertains to the design of *situated* systems, that is, of systems that contain not only abstract processing elements, but also sensing and actuation elements. This leads us to categorize components as *sensing components (S), processing components (P),* or *actuation components (A)*.
- The second extension pertains to the inclusion of *interchangeable human as well as machine* components. This leads to categorize specific instances of a component as being of *human nature (h)* or of *machine nature (e)*.[1]

Thus, in total we have six possible combinations: Sh, Se, Ph, Pe, Ah, and Ae.

A *sensing component* transduces information from the physical world to its output and does not receive any informational input from the system. Examples of sensing components are: Electronic Cameras (Se), Electronic Temperature Sensors (Se), Humans providing textual reports of what they have seen (Sh), Humans providing reports of whether they heard a particular sound (Sh) and so on.

A *processing component* transforms information from its input to its output. For example, one could have an electronic Face Recognition processing component (Pe); its input is a stream of images of faces, its output is a list of estimated identities. Or one could have a human processing component (Ph) that provides a natural language-translation service; its input is text in English, its output text in Chinese.

Finally, one can also have *actuation components*, transducing information from the input to action in the physical world; for example, a robotic arm (Ae), a light display (Ae), or a human which is capable of picking up and placing objects (Ah), or singing songs whose scores are sent to him electronically (Ah). The pictorial representations of the three basic components are shown in Figure 1, together with the connector links that connect them in a processing pipeline.

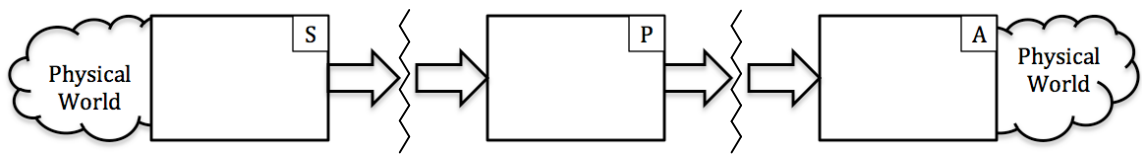

**Figure 1**: The three basic component types: S (Sensing), P (Processing), A (Actuation)

---

[1] The distinction between components and component instances should be noted at this point: a component is the class of the individual pieces of hardware or people (component instances) that perform the exact same function and are perfectly interchangeable.

For the case of human components (Sh, Ph, or Ah), a special human-machine interface is required, which is internal to the component itself. This might be their cell phone, a touch-screen interface; a keyboard; a spoken-language dialogue system, a browser application, an online game, and so on. Essentially, for the case of human components, what we have is something similar to Figure 2.

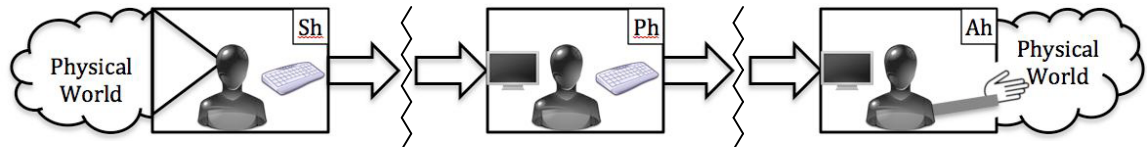

**Figure 2**: Internals of the Human versions of the three basic components: Sh, Ph, Ah. The Human-Machine Interface is pictorially depicted (symbolized by keyboard and screen), inside the wrapper of the component.

In Figure 2 there is also a depiction of generic input and output devices for human-machine interfacing specific for the human components. Research questions and implementation tasks for human-machine interfaces suitable for human crowd-servicing, belong to L0 (Real-time Human Services Interfacing) of the supporting mechanisms for CLIC. Also, research questions regarding incentivization of humans belong to L0 too.

An extensible yet compact *typology* of possible sensing, processing, and actuation components, either human or machine, is one of the features of CLIC. Such a typology should also support multiple levels of specification abstraction, as we shall see: moving from device-specific towards device-independent descriptions. This typology is developed as part of Layer L1 (the Cognitive Component Interfacing).

It is worth noting that each of the above components comes together with an associated *service-level agreement (SLA)*, prescribing its supported functionality, quality, pricing and so on.

Components which are available for use belong to the *Component Pool*. They have to declare their availability through the mechanisms provided in Layer L1 of CLIC, the Cognitive Component Interfacing Layer. This Layer prescribes SLA structures, as well as "wrappers" for making specific components able to be interfaced with CLIC.

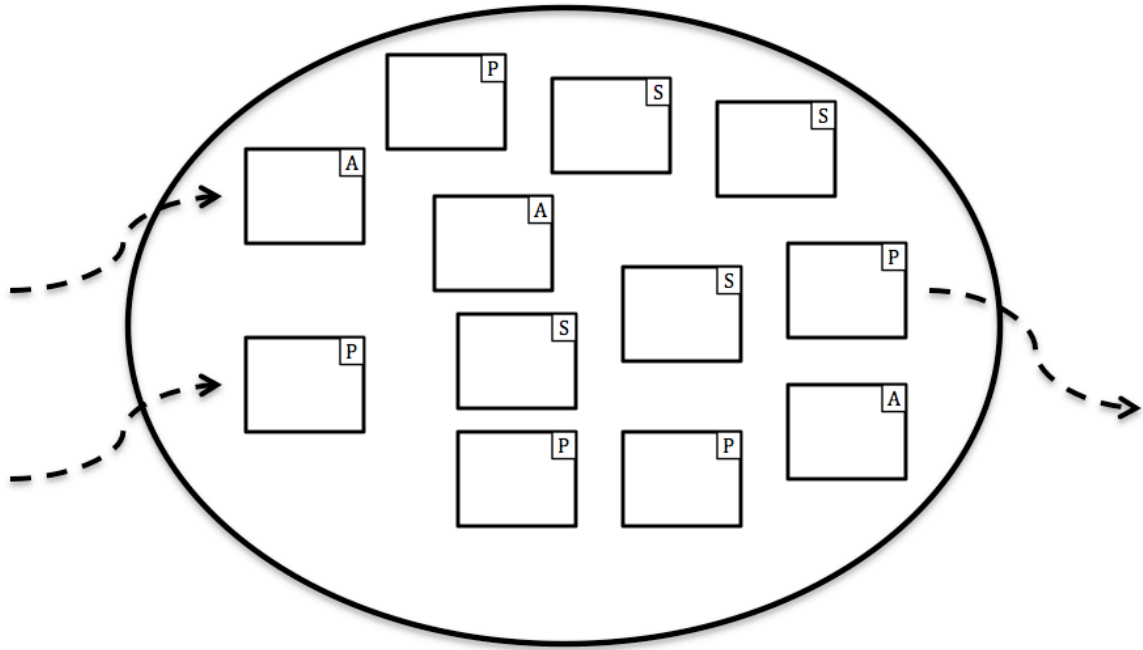

**Figure 3**: Component Pool: where available components reside, while other components exit or enter

So far, we have just talked about basic components, and several aspects of them: notation, types, special provisions for human components, and mentioned the lowest layers: L0 and L1. We have also introduced the Component Pool as well as the Information Pipeline. The important question that follows is: how can one assemble components and bind them in order to form an information pipeline, which will implement a Cognitive System within CLIC?

CLIC provides two main avenues for doing that: one can start by giving to CLIC either a mid-level *"Blueprint Specification"* of the desired Cognitive System, or a high-level *"Teleological Specification".* Let us start with the first. A Blueprint Specification provides a description of the information processing pipeline that will implement the desired cognitive agent: how many components, their overall type, as well as the topology of their interconnection are specified in this blueprint. Also, each component has a partial specification regarding its functionality and characteristics. Thus, we do not have a full specification for each component that partakes in the pipeline; but just a partial specification, i.e. what we call a *blueprint*. For instance, the partial specification might declare that we need to have a sensor that can detect humans in a specific spatial area – for example, in front of the main entrance of a specific university. There might be many available sensing components in the component pool that might fulfill this partial specification; for example, a human with a PDA willing to provide a report; or a street camera; or even a UAV-mounted camera that can fly there within the maximum timeframe provided by the specification. All of these components that are available in the pool could potentially *implement the blueprint*. The search for available components that could implement a blueprint, the negotiation of their SLA and their reservation, as well as the monitoring of their operation in order to ensure they are not faulty and that they conform to the SLA, is taken care of through the mechanisms of *Layer L2*, usually referred to as the "Service Procurement and SLA Monitoring" Layer. Furthermore, L2 initiates information flow through the pipeline, once it has been implemented and its components have been bound.

The second avenue, which is to give to the system a high-level teleological specification, is handled by the top layer (L3) of the CLIC architecture. This layer will also deal with possible conflicting goals between the actors of the system. The details of how this layer and all the other layer work will be detailed in the next section.

## 3. Architecture

We propose a four-layer architecture, which provides mechanisms for: Goal Arbitration and Teleological Description at the highest level (L3), Service Procurement and Replacement and SLA Monitoring (L2), and Cognitive Component Interfacing (L1). The low-level mechanics of indexing and connecting the available components is treated in pseudo-layer L0 that, although essential, is outside the core cognitive architecture, since they provide the infrastructure of using and negotiating with the offered services, and form the cloud of the machine and human services that the cognitive system will be made of, but they are not part of the CLIC framework itself, which is the way that we select which of these services are going to be contracted.

Let us now illustrate the above functionality through a concrete example. Let us assume that a user of CLIC desires to create a simple Cognitive System that is able to make a loud noise inside a certain office when a specific person stands in front of the entrance of a certain building. The user provides Layer 2 with the following "Blueprint Specification": (Notice here that for the sake of simplicity we have chosen a basic three-component feed-forward system. Real world systems implemented through CLIC could potentially have thousands of components with complicated interconnection topologies)

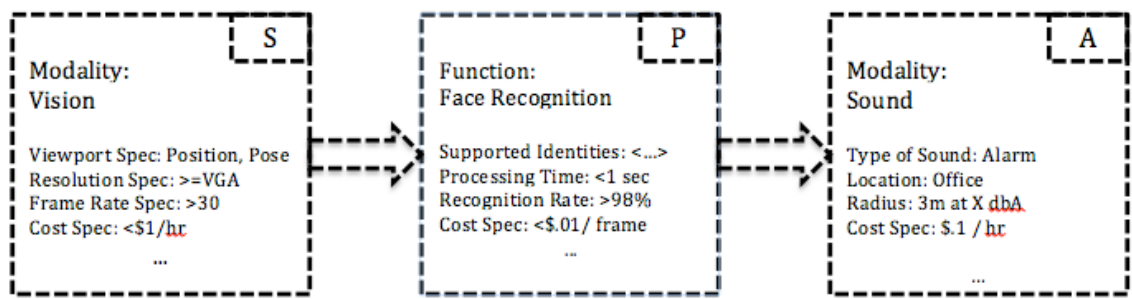

**Figure 4**: Example Blueprint Specification of a basic three-component cognitive system

Once the Blueprint Specification has arrived at Layer 2, the mechanisms of the layer search the available Component Pool through the interfaces of Layer 1, in order to find available components which fulfill the partial specification. A number of actual component candidates arise for the implementation of each of the blueprint components. The SLAs of each of the candidates are fetched; and possibly after negotiation, one specific available component is selected in order to implement each blueprint component.

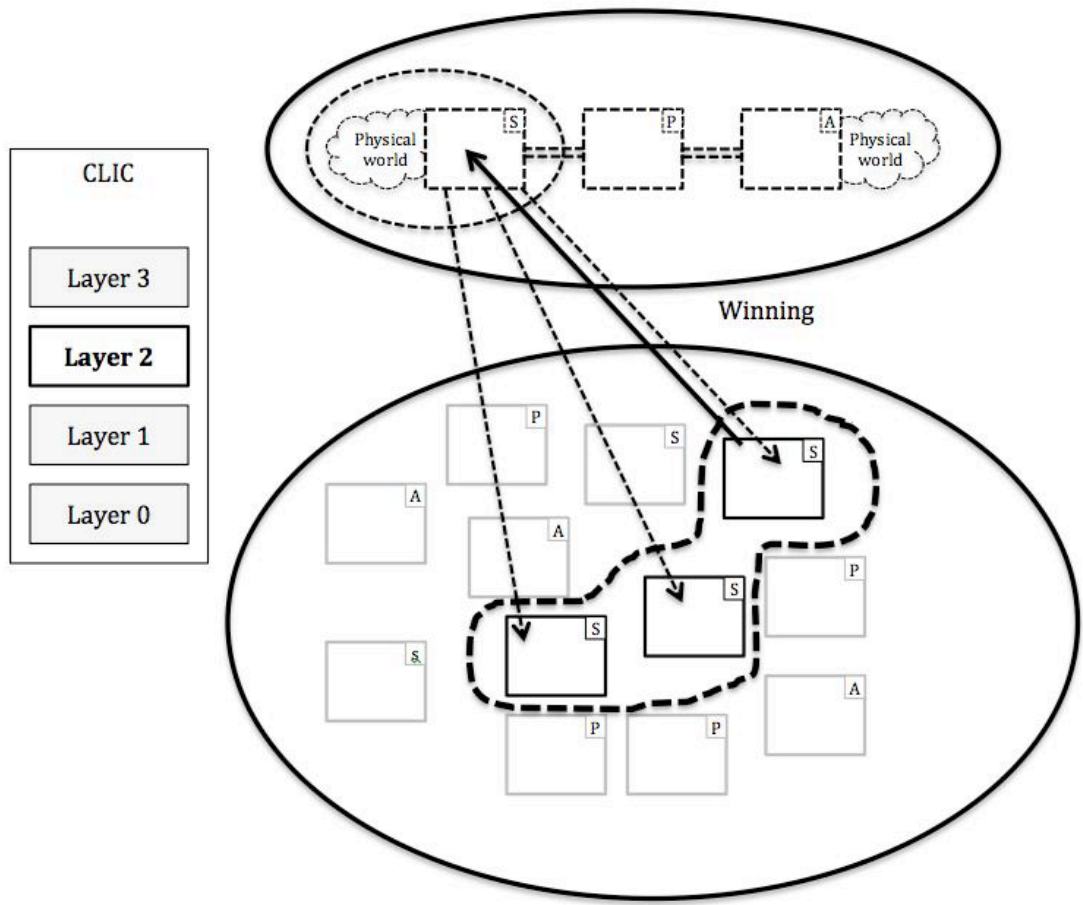

**Figure 5**: Searching the available component pool, examining suitable candidates, and selecting the winning component in order to implement the first component of a Blueprint Specification

Once the winning components are selected, they are reserved through a contract. Then, they are interconnected through segments of the information pipeline (Layer 2 utilizes the interfaces of Layer 1 towards that goal). Still, information flow has not started. According to the starting time specified in the Blueprint Spec, information flow is initiated by appropriate signaling from Layer 2 to the components.

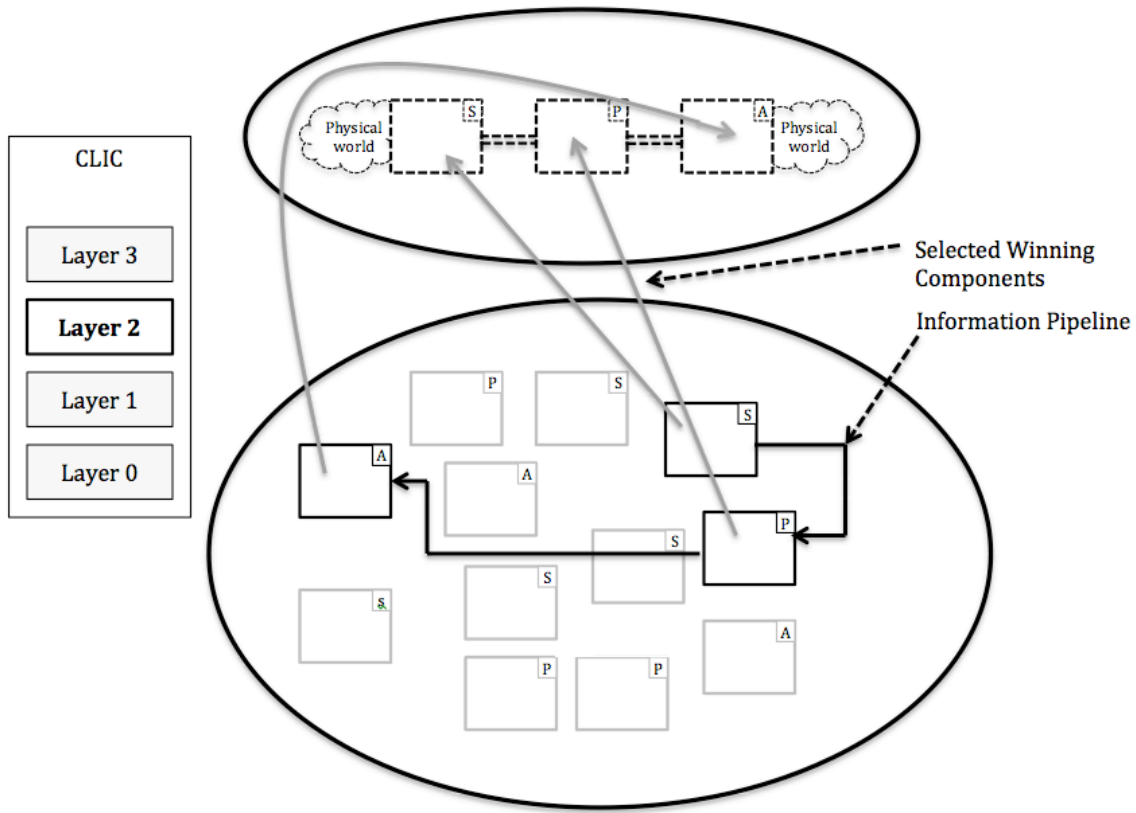

**Figure 6**: Interconnecting the chosen components to the information pipeline and starting the information flow.

After information flow is started, the user's desired cognitive system which was described in the blueprint specification has been actually implemented and is furthermore in operation. During operation, Layer 2 has the important role of monitoring operation, and replacing component when required. Component replacements may be required if a component unexpectedly becomes faulty or unavailable; if the contract end time is reached; if the terms of the SLA are not kept; or if the availability of a better or more economical component makes the replacement advantageous, despite any potential penalties.

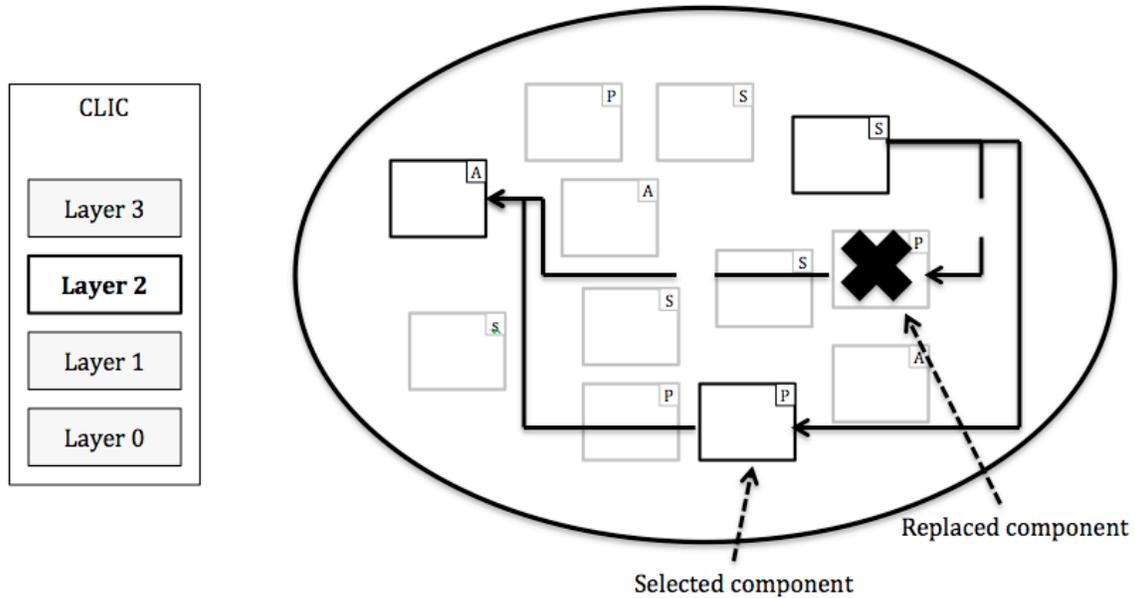

**Figure 7**: Example of replacement of components

So far, we have shown how components can be selected, and rented and replaced, in order to implement and operate a partially-specified desired Cognitive System. Through the mechanisms described above, one is able through CLIC to create Cognitive Systems that are situated, whose components can be human as well as machine, which exhibit high robustness, and which can be constructed on demand as well as reconstructed on the fly. However, we haven't yet illustrated how the components of such a system can be not only reused, but also time-shared. Reuse can easily be envisioned; once the operation of a certain cognitive system is terminated, they are free to be procured by Layer 2 in order to participate in another Cognitive System, which might have been requested to be created and operated through CLIC.

CLIC can reach even higher utilization of components and minimization of their idle time. This is illustrated in the following scenario. While the person-detection-alarm Cognitive System described above is still in operation, a blueprint specification for a new Cognitive System that has to operate at the same time is sent to CLIC's Layer 2. While the previous cognitive system is operating, tis components might be time-shared as illustrated in the following two scenarios. First, a component might be able to either *share its output* with another recipient (for example, the output of a camera might also be fed to a car detector, in parallel to the face recognizer; in that respect, the Sensing Component S can partake in both a cognitive system performing human detection as well as in another which does traffic estimation and reshaping). Second, a component might have unused processing cycles; these could be reused, by it partaking in another cognitive system which requests its services, through *time-sharing*. For example, there might be another cognitive system which is performing massive tagging of faces in a video that was shot some months ago; that system might be able to utilize unused cycles of the face recognition Processing

Component P, which is also partaking in the person-detector alarm cognitive system described earlier in this section.

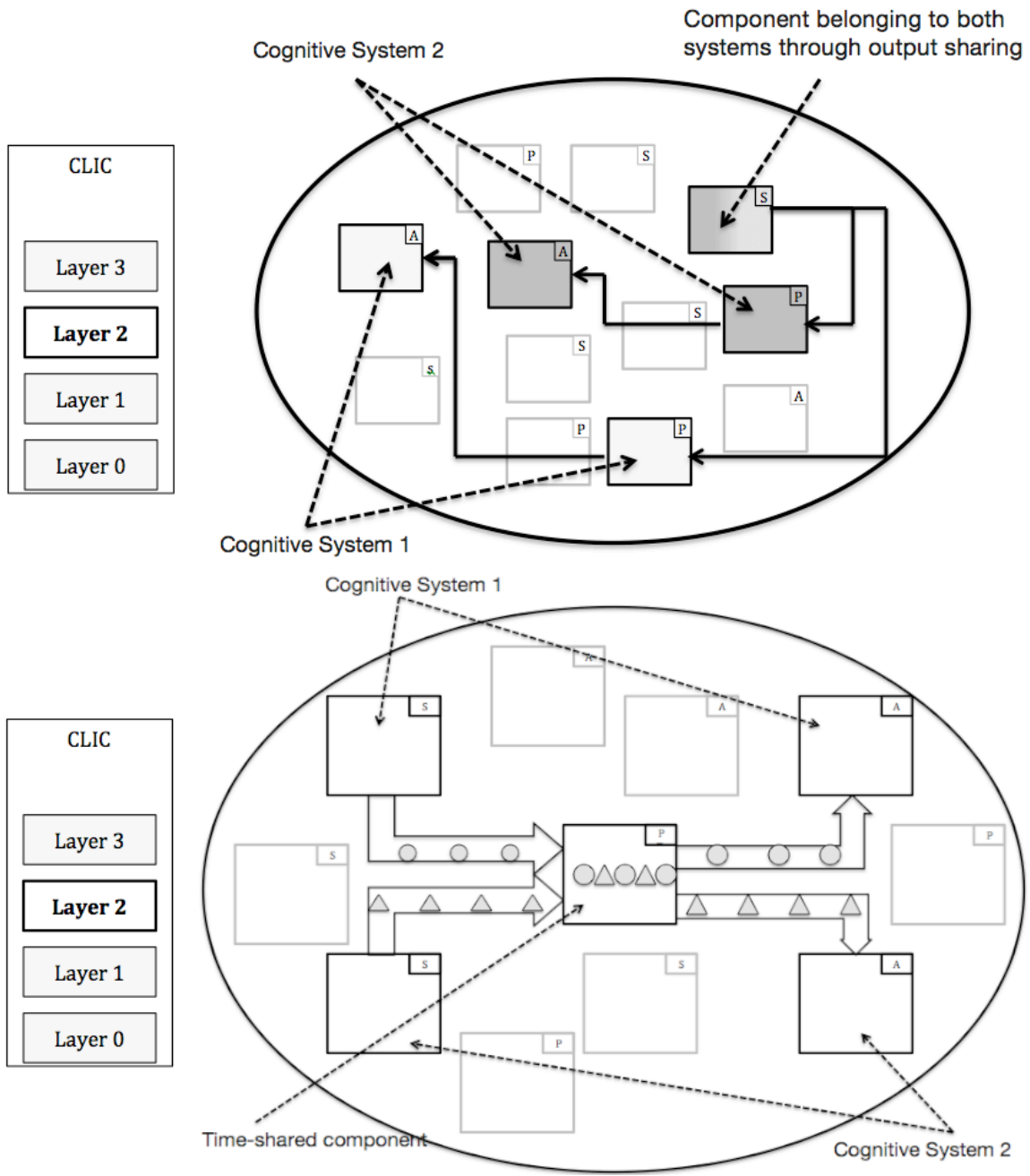

**Figure 8**: Two ways of re-using components: output-sharing (top) and time-sharing (bottom)

Finally, having described the involvement of the mechanisms of Layers 0, 1 and 2 in the operation of CLIC and the construction, operation, and maintenance of desired Cognitive Systems through CLIC, let us describe the main functionalities of Layer 3, which has two main functions: *Goal Arbitration* and *Teleological Description*

*Translation*. Goal Arbitration takes place during the run-time of a Cognitive System deployed through CLIC; Teleological Translation takes place before the construction of a Cognitive System through CLIC, if desired.

Let us start with the latter: *Teleological Description Translation*. As mentioned above, there are two ways for a user of CLIC to request the creation of a particular Cognitive System: through a Blueprint Specification, which is sent to Layer 2, or through a Teleological Specification. The Teleological Specification does not contain a description of the structure of a system and the specifications of its components, as is the case for the mid-level Blueprint Specification; rather, it contains a high-level description of the desired purpose of the system, and not of its structure. Layer 3 provides tools for creating such descriptions, and then, for translating them to mid-level Blueprint descriptions, which can be sent to Layer 2 for procurement and operation, i.e. effectively aiding in the translation of high-level teleological descriptions to mid-level structural.

Now, let us move to *Goal Arbitration*. While a certain cognitive system is in operation, one might able to externally dynamically specify parameters of the current goal that it is trying to pursue. For example, in a traffic control scenario, the corresponding system would have to serve two kinds of goals: (a) preset ones, such as reducing traffic congestion as well as pollution, and (b) dynamic ones, such as maximizing the satisfaction of the drivers who have declared their own desired destination points, by reducing transit time etc. The total goal of the system can be viewed as a composite of all the above dynamic goals as well as its preset goals. Many of the component goals are antagonistic, or even worse, might be mutually impossible. Thus, the purpose of the goal arbitrator is to be able to create a satisfactory composite goal, in view of these potential discrepancies.

Last but not least, after this more detailed conceptual introduction to CLIC, and its layers of mechanisms, as well as of the Cognitive System specifications and its creation and operation processes, it is worth commenting upon one of the main resulting changes catalyzed by CLIC: Through CLIC, the creation and operation of Cognitive Systems is *transformed to a utility provided on-demand*; which would lead to an increase of component utilization and thus potentially to *large economic consequences*. For example, one does not need to own extensive sensor networks (for example cameras in wide areas) in order to be able to operate services using them; they can be time-shared and procured on demand, when and if they are needed. The same holds for processing and actuation services; in this way, an *open market* of component services is created. We are thus envisioning a world where processing code can run in machines in various locations, expert and non-expert humans provide crowd-servicing in time-slices of seconds, and furthermore robots and sensor networks can be freely and transparently procured and interconnected in order to create the Cognitive Systems of the future, through CLIC. In essence, *Situated Intelligence is transformed to a commodity*.

Finally, apart from the economic and open-market far-reaching consequences of CLIC, there is yet another option arising through it: The humans providing their services to CLIC, by implementing sensing, processing, or actuation components for it, can choose to do so for a variety of possible *incentives*: for monetary reward, for the sake of a desired common cause, as a volunteer, or in exchange for having CLIC itself fulfill their requests; thus, the option for partial *auto-telicity* arises, as the goal of the cognitive systems that are created can arise from within them – and one can explore advanced ways of mediated participative collective decision making, as will be illustrated in our traffic control use case presented in the next section.

## 4. An Example Use Case: A Traffic Controller Cognitive System

We will illustrate the concept of CLIC systems through a realistic yet demanding real-world example: An innovative traffic management system construed as a *single cognitive system* containing *humans* (e.g. drivers, pedestrians, or controllers) as well as *machines* (e.g. software, cameras, or traffic lights) participating in the cognitive system. More specifically, *Sensing* for this huge cognitive system will be provided by human as well as machine components (e.g. human verbal reports, traffic cameras, or GPS), *Processing* will be provided by humans as well as machines, and *Actuation* will be provided by human (car driving) as well as machine (traffic lights, announcement boards) components too. Also, it is worth noticing how the *purpose* (teleology) of this *situated and distributed cognitive system* will arise: The purpose of this system is constructed as a negotiated mix of the purposes of the drivers (desired destinations and route characteristics) as well as of global goals (traffic optimization), in essence creating a *partially auto-telic* system. Furthermore, the system is highly fault-tolerant, and directly scalable to different and larger cities.

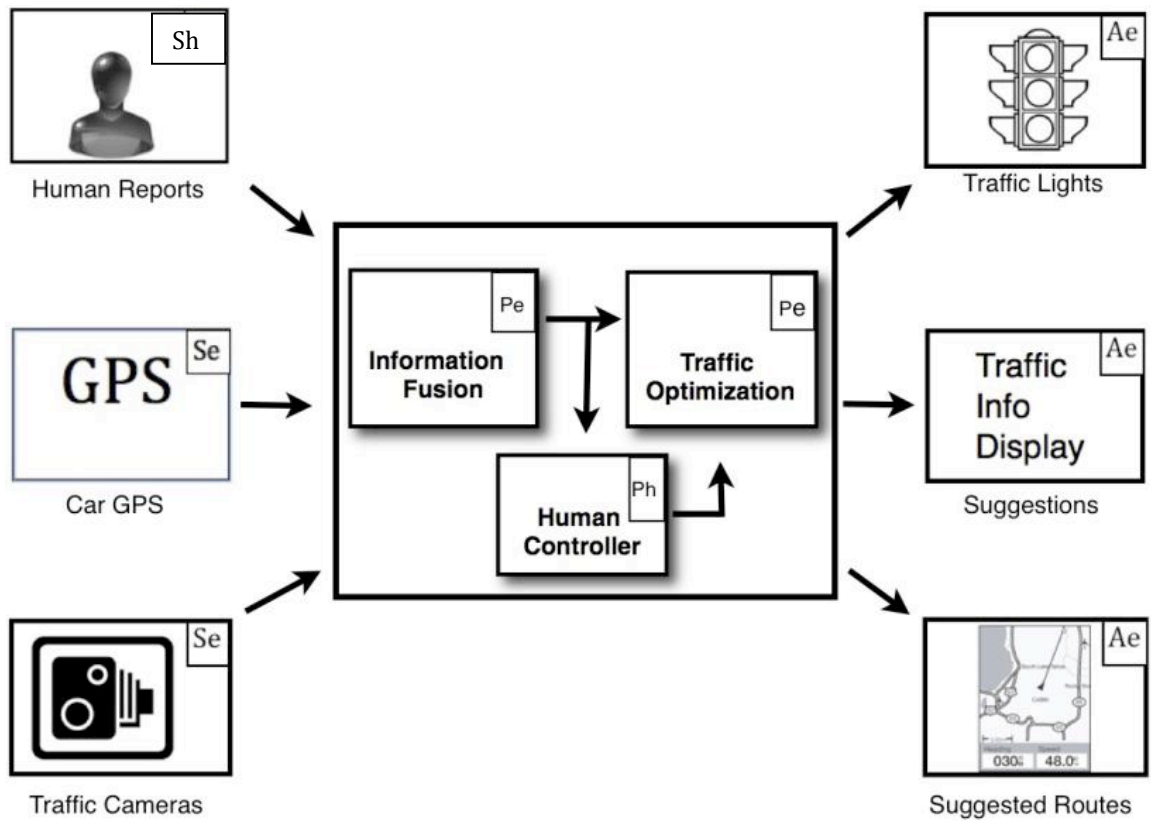

**Figure 9**: Block Diagram of example Use Case: Human-Machine Cognitive System for Traffic Control

In this giant unitary cognitive system, the following dynamic components partake (see Figure 9):

*Sensing services* are provided by human as well as machine components. Human traffic reports are acquired through driver-interfaces, either whenever the driver wants to report an important event or when the system proactively asks a specific question to him. At the same time, traffic cameras, whose output is reusable for other services too, provide traffic data; and the GPS units of the car-installed PDAs of drivers participating in the system provide real-time position information for a subset of the vehicles of the city.

*Processing services* are provided by humans as well as machines: Sensor fusion, goal arbitration and traffic optimization services take place here, with special GUI Interfaces for the human controllers (this will be described in the next chapter).

*Actuation services* are also provided by both humans as well as machines. Humans receive suggested routes, which they are incentivized to follow; also, human traffic police regulate important points of traffic if required. On the machine side, traffic lights are controlled; as well as announcement boards, which are readable both by the drivers that participate in the project and have online driver PDAs, as well as those drivers that do not, and which can indirectly modulate flow through announcements of estimates and suggestions. Of course, this modulation takes place at a macro-level; (many cars together) and not at the micro-level (each car - separately), as is the case for the drivers participating in the system through their dedicated PDAs.

But any cognitive system, apart from Sensing, Processing, and Actuation, requires a Goal. It is worth noticing how the *purpose* (teleology) of this *situated and distributed cognitive system* will arise: The purpose of this system is constructed as a negotiated mix of the purposes of the drivers (desired destinations and route characteristics) as well as of global goals (traffic optimization). The system is essentially trying both to satisfy the goals of the individual drivers, as well as to keep the traffic moving and the environment of the city in a good condition.

The use case, apart from being a real-world scalable problem of importance on its own right, acts as a very good illustration of the concepts and mechanisms of CLIC, which could easily be applied to a huge number of other scenarios, including disaster response, law enforcement, surveillance, military, as well as medical and postal process applications, assistance of the elderly or citizens with special needs, and many more, where fault-tolerant distributed human-machine systems are required.

## 5. Research Directions

In this section we will give more details of the four layers of mechanisms of CLIC, and present associated research motivating the construction of CLIC in related fields. We will start by discussing the lowest-level layer L0, and then proceed all the way to the highest-level L3.

### *5.1 L0: Interfaces for real-time human services*

The aim of CLIC is to procure and process data from a large number of people; data that has to be related to or combined with data and services procured from machines. It is up to L2 to contract the users in *the best way possible*. Requesting information from people needs to be done through a user interface (UI) accessible at this lowest level, whose goal is to collect appropriate data from the humans acting as service providers. This data will contribute to new types of services since it needs to be combined with data from human sources or from both human and machine sources (described previously and illustrated in Figure 6 and 7). These services, in turn, will be used as input to the adaptive service procurement layer (L2), e.g. the next layer in the architecture (described previously and illustrated in Figure 8). This UI should not be confused with the user interfaces handling high-level cognitive enquires or tasks (at L3).

Using data from different collocated or distributed assemblies of people and processing is actively investigated by various HCI related areas, e.g. social psychology, cognitive sciences, behavioral sciences or sociology. The interfaces used to procure data from humans need to be kept simple and accessible through several different appliances. Since the focus is to show possibilities of providing services by combining human and technology sources, the starting point is to use existing UI design solutions associated to data processing methodologies considering humans as data providers e.g. from crowd-centric technologies [Kit08], from mashup studies [Hol09], or by using on-line communities (e.g. for emergency and helping public safety agencies [Pal11]). The departure will be based on methods used from crowd-sourcing, examining ad-hoc online communications or mashop studies.

Regarding existing work in crowdsourcing, a prime example of real-world platforms is the Amazon Mechanical Turk (MTurk), one of the suites of Amazon Web Services, that enables computer programmers (Requesters) to co-ordinate the use of human intelligence (Workers) to perform tasks that computers are unable to do yet. These services have been used already for a wide variety of tasks, including speech transcription, translation as well as evaluating translation quality [Cal09], annotation [SorFor08], user studies [Kit08], and much more. The pioneering work of Luis Von Ahn on Human Computation also offers a good illustration of real-world crowdsourcing, for tasks such as labeling towards automated object recognition [AhnDab04], as well as on exploring the potential of game-like activities for harvesting human cycles towards collectively solving large-scale problems [Ahn06]. Of particular relevance to *CLIC* and in particular to the concept of the Human-Machine Cloud are a number of special issues, such as appropriate task routing to people with relevant abilities, which can be implemented for the case of relying on local-knowledge of the agents involved through the approach of [Zha11], ingenious approaches towards achieving simple but effective quality control such as the iterative dual-pathway structure used for speech transcription in [Lie11], and consideration regarding high response speed real-time crowd-powered systems [Ber11].

The main novel characteristic is the need to offer a real-time service, so that the UI can, for example, decide if this is the right moment to interrupt the user with a service request or not, and the UI must provide support tools for honoring the SLA. These would include e.g. a countdown timer, so that the service provider knows if it's still worth to complete the task or she is already in violation of the SLA and will not get paid anyway.

Based on earlier studies on e.g. considering patterns from social and technological sources (e.g. [Sch06], [Hel07]), to process information this part of the project will be able to contribute towards understanding collaboration in complex situations. These will open up research questions, such as questions on social possibilities in technical settings, for example to define usability requirements for treating data from combined sources. An overall idea about the UI needed here (in relation to the main parts of this project and the overall data flow is in Figure 10).

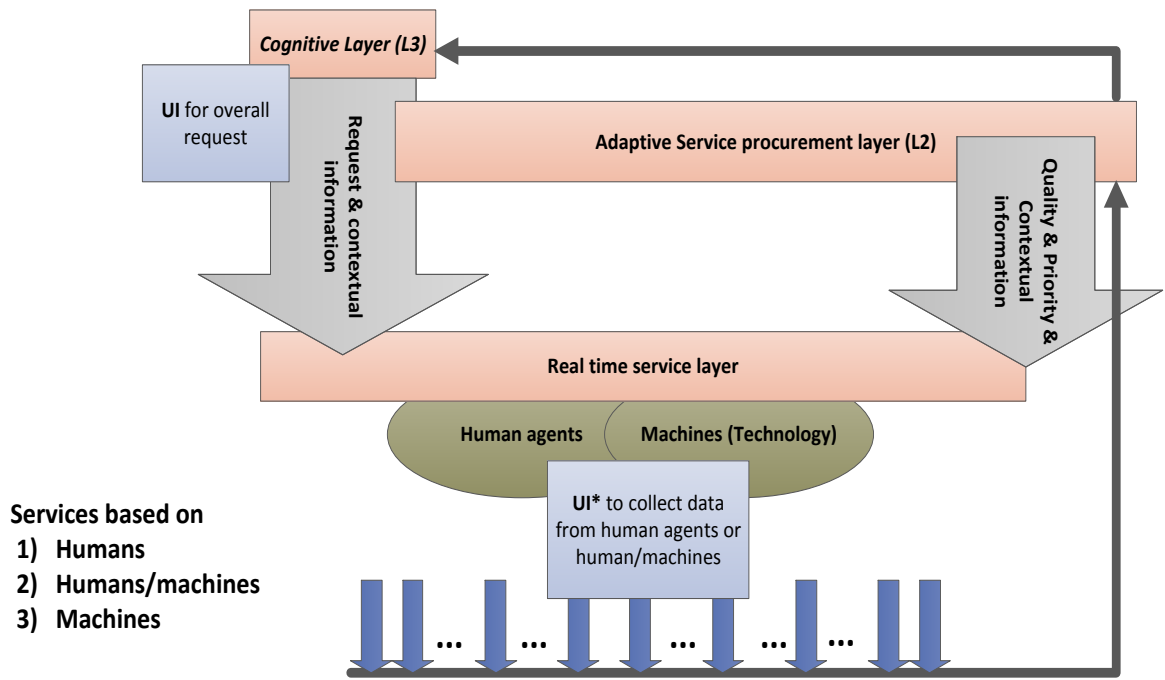

**Figure 10:** Behind the **UI**\* here is an overall request (given in L3) with contextual information. This together with automatic information from L2 has to be prepared so the human agents easily can contribute with data. The human agents need to fill in their data via **UI\***. This request has to be meaningful (request for services from humans or from humans/machines – if the machines services are not enough), through the **UI\***.

To start with, we can begin to collect data from human agents with help of the commonly used Application Programming Interfaces (APIs) e.g. through Java and PHP (to use internet oriented languages, instead of trying to decode natural languages) [Kre10]. Software development APIs make it possible to create new applications requiring no permissions and little programming knowledge. There are possibilities in e.g. identifying keyword patterns – for aggregating data [Bol05]. These keyword patterns can be considered identifiable by knowing the context and also given the practical pilot project that this project can follow from the beginning. Given the increasing number of humans engaged in micro-blogging [Mil08] together with the possibility to use mobile interfaces, this project can engage human agents to contribute the needed data. Also there are technical possibilities to access human information – with possible adjustments on actual interfacing and, in the beginning, using existing web applications. Using patterns to contribute to better interfaces can be further developed, especially after examining the data obtained. This will contribute to the above mentioned *crowdservicing*. There are also methods that improve this by following the information, e.g. using the method suggested by Kittur for crowdsourcing [Kit12].

Our research plan is to investigate how social and technical information can be visually represented and followed through these complex settings, contributing to research on universal interfaces. By testing new concepts via defining prototypes for interfaces that consider data from mobiles and computers, from social actors, sensors and machines this project contributes to better understanding universal interfaces that satisfy unobtrusiveness requirements, while at the same exposing APIs that uniformly apply to human-provided and machine-provided real-time services.

## 5.2 L1: Cognitive component interfacing

In order to materialize the CLIC framework, we borrow and combine architecture and integration solution from two different communities: from *Service-oriented Architectures (SOA)* and IT services infrastructure related to real-time services and service composition and from *robotics* and *cloud computing* infrastructure related to integrating heterogeneous components, including *Sensing, Processing* and *Actuation*.

Services in general, including Web services in IT, services in business and governance, or any other part of the services system, have received considerable attention from both academia and industry; culminating to a considerable volume of literature and specifications that address the description, discovery, and composition of services.

Focusing on the IT perspective, different standardization bodies have produced different specifications for *Service-oriented Architectures (SOA)*. The World Wide Web Consortium (W3C) produced the *Web Services Description Language (WSDL)*, a recommendation for specifying the format and values of data flowing into and out of Web services [WSDL2]. In other words, WSDL provides a standard XML schema language for specifying the *syntactic* details of how to invoke a Web service; the *semantics* of these services and data can be expressed by annotating WSDL components with identifiers from a semantic model [SAWSDL]. Similarly, the Object Management Group have recently finalized the *Service-oriented Architecture Modeling Language (SoaML)*, a meta-model for the specification and design of services within a service-oriented architecture [SoaML]. SoaML is more geared towards using UML as a modeling language instead of Semantic Web technologies used by W3C, but other than that both specify SOA. A recent trend in SOA is to provide RESTful services, that is, Web APIs over the HTTP protocol. The *Web Application Description Language (WADL)* aims to become the standard in describing RESTful services [WADL].

All these generic SOA frameworks specify the mechanisms of service invocation and data exchange but do not delve into the specifics of defining schemas and vocabularies for encoding the semantics of these actions. That is to say, specifying the semantics of, for example, a camera offered as a service would require vocabularies for specifying inputs (e.g., coordinates of the location the service consumer needs filmed), outputs (e.g., video encoding), and the service provided (e.g., real-time video from the input location). Such vocabularies receive their semantics from *conceptualizations* or *reference models* that define the top-level concepts and relations of the domain. The *SOA Reference Model* [SoaRM], for instance, is a reference model for specifying services in SoaML. Using Semantic Web technologies, OWL-S aims to abstractly modeling services and their properties [OWL-S]; although in the OWL-S specification WSDL bindings are used as example groundings of OWL-S concepts, OWL-S can be used to describe any services infrastructure. Oberle *et al.* [OLG06] provide a solution with the DOLCE foundational ontology, a modular and extensible general-purpose top-level ontology [GGM02]. DOLCE has been often used as the foundations over which to develop service ontologies, more recently by Ferrario, Guarino, *et al* [FGK11] who propose a unified technical and business perspective of services.

The latter work shares with *CLIC* the vision of a unified abstract description of services provided by people (in, e.g., the context of business processes) and of services provided by machines (in, e.g., *software as a service* settings). What has never been explicitly foreseen previously is the provisioning of the same service, or rather of directly interchangeable services, by either humans or machines. Current models provide for the description of a service's properties, including its inputs and outputs and its service capacity; querying for an appropriate service to fill a position in a pipeline amounts to looking for matching inputs, outputs, processing, and setting *acceptable margins* for parameters such as the price and quality of the service or the speed or rate at which the service is provided.

What is missing is support for the complex situation arising from the involvement of human real-time services, delivered over an intelligent, unobtrusive user interface: in most situations it would be necessary to *negotiate over multiple issues* at the same time, including negotiating over attracting the service provider's attention in the first place. In other words, and regardless of the negotiation strategies (discussed in Section 5.3 below), it must be possible to represent the price offered and the quality of service expected by the service seeker in relation to each other and not as orthogonal parameters. Symmetrically, the unobtrusive and extendable UIs (discussed in Section 5.1 above) must be able to advertise not just the provision of services, but also (depending on the user's current activity) the minimal price/service level offer that may interrupt the current activity.

*5.2.2 A cloud of sensing, actuation, and processing services*
The need to integrate heterogeneous components is very strong in the *Ambient Intelligence* community [AHS01, SS08]. In Ambient Intelligence, there is a strong trend to build over the *Open Services Gateway initiative (OSGi)* framework, a platform for the Java programming language. OSGi foresees dynamic component addition and removal, as well as *publish-and-subscribe*, event-based inter-component communication. OSGi has been used as the basis for integrating sensing, actuation, and processing components in Ambient Intelligence [DVC08]. Similar frameworks have been developed in *robotics*, where many popular integration frameworks facilitate inter-component communication by providing middleware for the exchange of typed messages where components subscribe as consumers or producers. *RobotOS* is a prime, and very popular, example [QBN07], although this approach is ubiquitous in robotics.

Although very similar to the *CLIC* vision, such approaches lack the scalability and openness of *CLIC* where we foresee a global market of amalgamated human-machine services, radically heterogeneous and pushing beyond the confines of a single smart home or smart lab or robotic platform. For *CLIC*, the *infrastructure as a*

*service (IaaS)* paradigm is a more promising starting point, as it circumvents centralized middleware for data exchange and follows an approach where components expose interfaces through which to directly communicate.

Naturally IaaS operates at a very low layer, where the only service offered is a virtual machine; however IaaS software such as Eucalyptus implements considerable functionality needed in *CLIC*; a Eucalyptus cloud has a (possibly distributed) *cloud controller* which handles registering and deregistering physical hosts, virtual machines, and storage, changing properties of the virtual machines, and querying about existing resources. The cloud controller connects to the individual *node controllers* in each physical machine, to know if things are up and running and to apply changes. Furthermore, *elastic storage* is implemented as a *software as a service (SaaS)* facility where access to a (distributed, variable size) data store is provided over an API and not as direct access to a storage device; SaaS storage abstracts over the specifics of where the storage device is located or even how many storage devices contribute towards a virtual store.

These concepts are very similar to those of our cloud of cognitive components, so that *CLIC* will explore cloud computing infrastructure, and not conventional message-queuing middleware, as the basis for binding its SOA conceptualization. What is missing from the current cloud infrastructure is the development of SaaS components for sensing and actuation that apply the same principles as elastic storage: abstract data producers (sensors) and consumers (actuators) that provide a utility without exposing any implementation details.

### *5.2.3 Implementing the CLIC framework*

The objectives we need to pursue in order to implement the CLIC framework relate to conceptual as well as implementation issues. At the *conceptual* level, we will develop a unifying conceptual reference model for human and machine real-time services and the service-level agreements and policies they are offered under. Besides a top-level vocabulary for describing taxonomic hierarchies of data and service types, this conceptualization must support the specification of soft limits, price/service level negotiations, and similar situations that arise from the involvement of human service provision. One promising direction is to assume as a starting point SOA extensions of the DOLCE foundational ontology, exploiting DOLCE's ability to *qualify propositions and concepts with complex parameters.*

At the implementation level, we will ground this conceptualization with a binding mechanism for providing such services as a scalable, extendable, and decentralized cloud. This mechanism will build upon current *cloud computing* infrastructure in order to endow robotics middleware with the scalable and distributed nature that cloud computing offers and robotics middleware currently lacks. This will lead to APIs and data type vocabularies for cloud nodes, decentralized service registries, and extendibility mechanisms for accommodating future and unforeseen service and data types.

## 5.3 L2: Adaptive service procurement and fault tolerance

To enable the procurement of the necessary services at L2, in order to implement the workflow of services provided by the cognitive agent at L3, the resources/components (both sensors, actuators and humans) at L1 needs to declare their capacities and service level in service level agreements (SLA). This includes both definitions of the service as well as the service level and the promised performance (e.g. quality or time of completion). For human resources, additional description is needed, such as the commitment and reliability of the service provided. The service levels of resources are declared by the resources themselves, and in addition, human resources declare their SLA as a promised level of service provided. The Service Procurement Agent (SPA) monitors that SLAs are upheld, otherwise it needs to revise and re-negotiate the set of services contracted to implement the task. This means that it may stop using some failing or under-performing services, in favor of other more reliable ones, for completing that task. Once the properties of the offered services are declared and monitored this way, adaptive service procurement management at level L2 can plan, monitor and revise resources used in the task.

There are a number of issues that the SPA must tackle in order to complete this task:

### 5.3.1. Selecting the Best Services

The SPA uses a dynamic database about available services (and their properties) in order that, for each task to execute, the currently most optimal services are allocated for a task. Optimal in this case would be determined by their price and service level that they offer and whether a service satisfies the minimum quality set by the workflow of services provided by L3. The SPA must negotiate what services to assign to the task prior to execution, depending on the currently offered services. Once a service is contracted, the SPA reserves this service for the duration needed by the workflow; if services fail to perform according to what is agreed - this information is provided to the SPA by the QoS estimation module - the SPA will locate alternative services to replace the ones that failed. In this way, the needed quality is ensured during the duration of the task. If there are insufficient overall resources/services in the cloud for completing the task, the cognitive agent (L3) is notified so that the overall task (which is implemented by the workflow) can be aborted or an alternative workflow can be issued to the SPA.

In the literature, e.g. in [SPJ09], there is related work about how to contract a set of cloud services in order to satisfy a workflow of tasks and maintain a certain level of quality while minimizing expenses. However, in all this literature, it is assumed that the price offered for each service is fixed. On the other hand, nowadays auctions are emerging as a suitable mechanism for balancing supply and demand in cloud service procurement settings [BAV05, MT10], and they are starting to emerge in real-world cloud systems, such as Amazon's EC2 spot instances and the SpotCould platform. Allowing service prices to be determined by supply and demand also provides an incentive to the providers to improve their services in order to get a better price for it, while at the same time it ensures that providers can always sell their service without having to find the optimal price for it, as this is determined by the auction;

on the other hand, if they set a fixed price and this is too high then they would not sell the service, whereas if they set it low, then they lose some profitability, hence the benefit of using auctions. This means that much of this pre-existing work, as it assumes fixed prices (no negotiation), cannot be applied to the setting that we propose.

### *5.3.2. Negotiation with Human Providers*

If all properties of the services offered are specified beforehand as is the case with all machine services, then it is most efficient to sell them in auctions as we've already discussed. However, humans can offer at the same time a potentially infinite array of options regarding the quality of the service that they offer. Hence, for most human services it would be necessary to negotiate with them over the quality and the price for that quality. One of the problems that we need to tackle when we negotiate with humans automatically is the ability to negotiate over multiple issues at the same time [SFWJ04], for example price and quality of service can be two issues to be negotiated together. In this context we will focus on the development of agent negotiation strategies and tactics [LC10] instead of other parameters such as the negotiation protocol or the information state of agents. Strategies are computationally tractable functions that define the tactics to be used both at the beginning and during the course of negotiation. They are based on rules-of-thumb distilled from behavioral practice in human negotiation. Tactics, in turn, are functions that specify the short-term moves to be made at each point of negotiation. They are structured, directed, and driven by strategic considerations [LWJ01].

Negotiation tactics and strategies is not a new topic in multi-agent systems research [LMNC04]. We need to study strategies and tactics for alternating offers protocols [Ru85, WMV02] based on concessions [DTT08] and their recent implementations, for example see [BUMTS09] and [US09]. We will focus on both bilateral [FSJ98] and multilateral [RZMS09] models. One of the key problems in these settings is that tactics and strategies have not taken into consideration the interaction of agents with humans and especially the fact that the human will tire easily, if they are offered too many offers and counter-offers.

### *5.3.3. Estimation of the QoS*

In order to know whether a provider honors her agreements with the system, it is necessary to be able to estimate the quality of the service both beforehand, meaning when the SPA contracts the resources and at the time when the service is supposed to be offered. Both the total performance and the capacity of each resource can be estimated, or learnt from the historical performance of the SPA (as done in QoS-oriented, self-aware networks [GLN04]). For human resources, capacity can be estimated similarly based on historic performance, combined with the SLA agreed and the actual performance. Therefore QoS is the matter of balancing the needs of multiple and concurrent users here, with all the available resources of the cloud. The traditional QoS approach uses a quote-for-price approach to estimate the total current cost of resources needed for a task, where the task submitter decides whether the price is acceptable.

Now, to obtain the correct QoS estimations, a flexible and open data infrastructure is needed comparable or possibly related to inputs obtained from human agents. Since this project includes human agents as data and service providers these input collected are mainly subjective experiences, observations or mainly by humans observable data. Therefore, for this project not only the estimation of QoS is in the focus, but we will relay on investigating QoEs (Quality of Experiences) [ITU08] in the traffic context.

### *5.3.3 Implementing the CLIC framework*
In CLIC, we use our expertise in designing trading agents that can obtain a set of resources while maximizing their utility (i.e. getting the resources that best satisfy one's goals) while minimizing expenses to create a strategy for the SPA that allows the procurement of the best services, i.e. maximizes the agent's expected utility. We will start from our methodology [VS03, VJS07] and use our work on realistic auctions [VJ10]. In addition to bidding in auctions for services and the flexibility that our methodology allows in re-planning in case that a contracted service should fail to deliver, our work will need to be adapted to the particular demands of the auctions for services; we will begin from some recently completed initial work about bidding in a single auction used in service procurement scenarios [VSJ12], which will be extended to bidding in multiple auctions. Furthermore, our design of the SPA will also be augmented to handle the multi-issue negotiations with the human providers.

In the negotiations with humans, we will cater for selecting the right level of granularity for selecting offers and related timing of an offer as humans will be slower than software agents and less tolerant to negotiate with small increments. This integration of negotiation models and their associated components has been long argued for [HR92] but has yet to be studied in detail, especially when humans are equal participants in the negotiation. Our contribution will therefore be to address the problem of negotiating with humans and agents at the same time, being able to discriminate between the two, and integrate these for both humans and agents.

Regarding the QoS issue, as the estimation of QoS alone is not sufficient, we will also rely on investigating QoEs (Quality of Experiences) [ITU08], within the domain context of the traffic management. Experiences are going to be approached in the context, by a number of reoccurring possible social and technical patterns [HEL07]. By identifying the main patterns a list of requirements can be provided. This list can stand as the bases of prioritizing requirements (also both the social and technical ones). Afterwards, a short action list with that part of the requirements that can be handled by the human agents can be defined. The list will contain comparable prioritized requirements and can be considered as a start-list that can be extended later (or replaced by e.g. more complicated algorithms). Since it is defined by humans (both system users but also service providers) as important requirements these can be used for predicting QoEs supported with objective metrics. This can provide basic information for identifying and simulating meaningful activities, workflows in the test environment. Therefore the requests contribute to instructions and also provide measurable data, with the right format to the next level. Even the data is limited for this pilot traffic context, basic requirements for estimating QoS and QoE can be examined.

The limitation to the traffic scenario context is important, since objective estimation for the QoE is hard. We will use simple lists of requirements to begin with, and then we will continue later by applying more advanced databases [MAT09] and more advanced software tools for automatic measurements, e.g. the one defined by Moore [Moo10], with specific focus on QoS estimations defined by Pessemier [PESS11], and considering methodologies for estimating temporal workflows [XIA12].

## 5.4. L3: Composite Cognitive Agency

There are two main challenges for this layer. The first challenge (C1) is how to specify cognitive systems at a higher-level, for example a teleological (intentional) level where the particular components the system is composed of are not known in advance, but only their desired cumulative properties and behaviors. The second challenge (C2) is how to arbitrate multiple goals which might also be coming from within the cognitive system itself, thus enabling partial auto-telicity.

Both of these are key challenges, but at least the first one is not within easy reach from today's state of the art. In *CLIC*, we have carefully tried to decouple them from the operation of the rest of the system; and, we have also decoupled the mechanisms supporting C1 from those of C2, too. Thus, one could well have *CLIC* in full functionality, creating and operating distributed human-machine cognitive systems on-demand and with on-the-fly replacement, without satisfying C1 in Layer L3; and this can be indeed the case, as long as the specification of the desired system is given as a mid-level structural Blueprint Specification – which we envision to be the norm most of the time.

However, in the future, we also envision that we will be increasingly able to use higher-level specifications, such as Teleological Specifications, which will have to be transformed to mid-level Blueprint Specs through Layer L3. In order to address challenges C1 and C2, the mechanisms of Layer L3 are broken down to two independent groups: (L3a): Higher Level Specifications - From Teleology to Structure - Generation of Blueprint Specs and (L3b): Goal Management and Arbitration.

### 5.4.1 Cognitive System Specification and Construction

Regarding L3a, of particular importance are existing theories and techniques that are related to purposeful agents and planning, especially as extended to distributed cognitive systems. There is a long tradition of viewing human interaction with different types of technology in terms of "joint cognitive systems" or "distributed cognition" in the overlapping research areas of cognitive systems engineering (e.g. [HW83], [HW05]), cognitive science (e.g. [Hut95]), and human-computer interaction (e.g. [HHK00]). The underlying assumption is that in many cases cognition extends beyond what is going on inside the heads of individuals. Much of this research is concerned with distributed human-machine cognition in environments such as control rooms or cockpits (for an example see [NLSZ12]). Most research in the area of *Cognitive Systems & Robotics*, on the other hand, has been concerned with autonomous artificial cognitive systems and, in some cases, the interaction with humans or between multiple artificial systems. Most of this research is still guided by the traditional assumption that cognition takes place 'inside the head' (or the robot equivalent thereof). From this perspective, *CLIC* goes beyond the state of the art by importing the idea of distributed or joint cognitive systems from cognitive systems engineering/cognitive science/human-computer interaction into Cognitive Systems & Robotics research. It also goes beyond the state of the art in distributed/joint cognitive system by considering the case of not only composite, but also dynamically varying cognitive agency.

The proper recruitment of services to satisfy user needs in such distributed and dynamic systems can in principle be related to planning and executing actions in a single-body agent. In both cases, an overall goal exists ("track this car through the city"; "drink from a cup") and in both cases, these tasks decompose into a sequence of "action primitives" that are given by the embodiment (e.g. recruiting appropriate cameras and OCR devices to read license plates; generating a sequence of reach for cup->grasp->lift->bring-to-mouth).

The crucial difference is that the embodiment in the *CLIC* case is not constant. Indeed, although it is known at a given time what services are available (since all providers announce them), it is not necessarily known for a service when it will be available (since providers can appear and disappear outside of the control of the overall system). The composite cognitive agency thus requires novel considerations in the planning aspects; at the same time, previous research, for instance into the representation of motion primitives in primate brains and the chaining thereof (e.g. [FFGR05; CTZB10; TZ10; TSZ11]) is still applicable given the overall similarity discussed above. The state of the art in composite cognitive agency is therefore furthered by applying insights from human and primate action planning to this more technical field of research. Simultaneously, cognitive theories of action planning need to be extended to account for the composite and dynamic nature of the embodiment in *CLIC*; requiring additional progress.

Our approach also expands upon and interconnects with existing work on agents which are built using a symbolic approach. Following the BDI (Belief-Desires-Intntions) architecture [Bra99, RG01], a well-known example is the KGP (Knowledge-Goals-Plan) model, where agents are construed in terms of cognitive capabilities interpreted as extended logic programs [KMSST08]. These capabilities enable the agent to sense the environment, decide which goals are most relevant next, plan for these goals, reason temporally and react to changes in the environment by attempting to act in it. CLIC revisits the KGP model in order to (a) make it compatible with our fundamental approach, i.e. specify a meta-model that would enable us to create agents in a plug-and-play fashion from components that would represent cognitive capabilities, including those capabilities provided by humans and their associated interfaces; and (b) study the relation between the unitary agent architecture of *CLIC* and its underlying multi-agent network that provides the required components. To address the many issues of how complex agent functionality within a cognitive agent scales up CLIC relies on the notion of super-agents [Sta10]. The original conception of super-agents is that of virtual organizations of specialized role-based agents that attempt to govern a resource by delivering the functionality of a single monolithic agent in a more distributed fashion. In this context, CLIC also seeks to identify how the cognitive systems we envisage relate to existing work on virtual organizations in Grid computing [MST11], including how to link these ideas to sensor Grids [SKPB07].

Furthermore, towards L3a, from a practical design standpoint, tools and methodologies that enable the designer of a distributed human-machine cognitive system to gradually transform teleological specifications to blueprints are highly relevant. We envision generalizations of the concepts of design patterns, BDI like plan libraries, simulink-like macro-blocks, or extensions of graphical block languages such as NXT-G as starting points, which could also later be produced starting from teleological descriptions, enabling the designer to semi-automatically fill-in the details. Towards semi-automated or fully-automated solutions, highly relevant are techniques attacking the problem from an AI planning approach [McITCS02] following a declarative approach. Planning can be interpreted as a kind of problem solving, where an agent uses its beliefs about available services and their consequences, in order to identify a solution over an abstract set of possible plans.

The starting point of the CLIC approach is that it introduces into the cognitive agents mechanisms that handle plan libraries, pre-specified plans in the form of workflows. This approach can be extended with more sophisticated techniques based on planning from first principles [KKC10] and extends them when necessary by looking at services as capabilities [SLK07]. In this context, it is important to study how to construct the coordination primitives of workflows and develop mechanisms that allow parts of the workflow to be constructed either automatically or with the help of the agent (artificial or human) who specified the high-level goal. This will allow us to acquire clarifications about the context of use prior to producing a concrete but inconvenient way of achieving the high-level goal by interacting with the remaining layers.

### *5.4.2 Goal Management and Arbitration*
Moving on from L3a to L3b, Goal Management and Arbitration (L3b), is quite interesting in its own right. Here, some of the central questions are: How can one combine intrinsic with extrinsic motivations within such a system? In our use-case scenario, how can one keep drivers satisfied, when they might have competing goals, and also, how can one also try to satisfice more global goals regarding utilization and pollution? Research in cognitive science has also concerned itself at length with intrinsic motivations of agents; including for instance some fundamental drives or a fundamental desire to minimize energy expenditure (for a detailed review/discussion of the relation between motivation/emotion and goal-directed behavior see [LZ11]). Within *CLIC*, insights from such research can be applied in a novel way for this type of system to assist goal arbitration. Specifically, it may be viable to define intrinsic goals (such as "to keep traffic flowing") and assign energy costs to different goals or different ways to execute them in such a way that minimizing the energy expenditure of the system corresponds to avoiding conflicting goals.

From a practical implementation viewpoint, there are also issues regarding the informational interconnection of the goal manager (L3b) with its input sources, and its communication to specific components of the cognitive system which is in operation, in order to alter their parameters of operation so that they can serve the current goal mix. The complementarity with Layer L3a is worth noticing here; L3b deals with *run-time* goal management, while L3b with *design-time* teleology (goal) and other high-level descriptions.

### *5.4.3 Implementing the CLIC framework*

After a thorough interconnection and juxtaposition of cognitive agency as it is viewed within CLIC to existing theoretical approaches, we will investigate the notion of teleology, and its relation to design-time vs. run-time goals, auto-telicity, intrinsic vs. extrinsic motivations, and elaborate on the multiple gradations existing between teleological and structural descriptions for CLIC-like distributed cognitive systems. We will use the theoretical devices developed in order to create software and tools that implement Layer L3, i.e. provide functionality for transforming high-level teleological descriptions to mid-level structural blueprint specifications (L3a), and for arbitrating multiple goals and enabling partial auto-telicity (L3b).

## 6. Conclusion

Motivated by the multiple shortcomings of the traditional approach towards designing and building artificial cognitive systems, in this paper we have presented CLIC, a framework for constructing and maintaining distributed human-machine cognitive systems on demand. CLIC's software architecture consists of four layers of mechanisms, proving support for real-time human-service interfacing (L0), cognitive component interfacing (L1), adaptive service procurement and fault tolerance (L2), and composite cognitive agency at the highest level (L3). Cognitive systems that are created and maintained through CLIC fulfill the following Desiderata: First, they are distributed yet situated, interacting with the physical world though sensing and actuation services, and they are also combining services provided by humans as well as services implemented by machines. Second, they are made up of components that are time-shared and re-usable across systems. Third, they possess increased robustness through self-repair mechanisms. Fourth, they are constructed and reconstructed on the fly, with components that dynamically enter and exit the system, while the system is in operation, on the basis of availability, and pricing, and need. And quite importantly, fifth, the cognitive systems created and operated by CLIC do not need to be owned and can be provided on demand, as a utility – thus transforming human-machine situated intelligence to a service, and opening up numerous interesting research directions and application opportunities. We started this paper by providing an introduction, following by a system and mechanism description, and an example of a use-case. Then, we moved on to a more detailed description of the four layers and their relation to existing research, as well as the important new research directions that open up.

Summing up, CLIC enables the design and operation of cognitive systems to move much beyond the current state of the art, by allowing the creation of fault-adaptive complex cognitive systems for a wide variety of application scenarios, where adjusting and recovering from failures is critical, and where wide-area networks of sensors and actuators, human and machine, are required. Most importantly, through CLIC, situated intelligence is transformed to a service, resulting to significant economic consequences, and opening up the new era of robust distributed human-machine situated collective intelligence on demand.

Most importantly, apart from this specific use case example, the mechanisms of CLIC enable the creation of other such fault-adaptive complex cognitive systems for a wide variety of application scenarios, where adjusting and recovering from failures is critical, and where wide-area networks of sensors and actuators, human and machine, are required.